# Optimizing Ensemble Weights and Hyperparameters of Machine Learning Models for Regression Problems

Mohsen Shahhosseini[1*], Guiping Hu[1], Hieu Pham[1]

[1] Department of Industrial and Manufacturing Systems Engineering, Iowa State University, Ames, Iowa, 50011, USA
* Corresponding author: E-mail: mohsen@iastate.edu

# Abstract

Aggregating multiple learners through an ensemble of models aim to make better predictions by capturing the underlying distribution of the data more accurately. Different ensembling methods, such as bagging, boosting, and stacking/blending, have been studied and adopted extensively in research and practice. While bagging and boosting focus more on reducing variance and bias, respectively, stacking approaches target both by finding the optimal way to combine base learners. In stacking with the weighted average, ensembles are created from weighted averages of multiple base learners. It is known that tuning hyperparameters of each base learner inside the ensemble weight optimization process can produce better performing ensembles. To this end, an optimization-based nested algorithm that considers tuning hyperparameters as well as finding the optimal weights to combine ensembles (Generalized Weighted Ensemble with Internally Tuned Hyperparameters (GEM-ITH)) is designed. Besides, Bayesian search was used to speed-up the optimizing process and a heuristic was implemented to generate diverse and well-performing base learners. The algorithm is shown to be generalizable to real data sets through analyses with ten publicly available data sets.

# 1. Introduction

Many predictions can be based on a single model such as a single decision tree, but there is strong evidence that a single model can be outperformed by an ensemble of models, that is, a collection of individual models that can be combined to reduce bias, variance, or both (Dietterich 2000). A single model is unlikely to capture the entire underlying structure of the data to achieve optimal predictions. This is where integrating multiple models can improve prediction accuracy significantly. By aggregating multiple base learners (individual models), more information can be captured on the underlying structure of the data (Brown et al. 2005). The popularity of ensemble modeling can be seen in various practical applications such as the Netflix Prize, the data mining world cup, and Kaggle competitions (Töscher and Jahrer 2008; Niculescu-Mizil et al. 2009; Koren 2009; Yu et al. 2010; Taieb and Hyndman 2014; Hoch 2015; Sutton et al. 2018; Kechyn et al. 2018; Khaki and Khalilzadeh 2019; Khaki and Wang 2019; Peykani and Mohammadi 2020).

Although ensembling models in data analytics are well-motivated, not all ensembles are created equal. Specifically, different types of ensembling include bagging, boosting, and stacking/blending (Breiman 1996a; Freund 1995; Wolpert 1992). Bagging forms an ensemble with sampling from training data with replacement (bootstrap) and averaging or voting over class labels (Breiman 1996a); boosting constructs ensemble by combining weak learners with the expectation that subsequent models would compensate for errors made by earlier models (Brown 2017); and stacking takes the output of the base learners on the training data and applies another learning algorithm on them to predict the response values (Large et al. 2019). Each method has its strengths and weaknesses. Bagging tends to reduce variance more than bias and does not work well with relatively simple models; boosting aims at reducing bias and variance by sequentially combining weak learners but is sensitive to noisy data and outliers and is susceptible of overfitting; while stacking tries to reduce variance and bias, that is, to fix the errors that base learners made by fitting one or more meta-models on the predictions made by base learners (Brown 2017; Large et al. 2019). In this study, we focus on stacking with weighted average as the second level learner, in which based learners are integrated with a weighted average. Although seemingly straightforward, the procedure for creating an ensemble is a scientific process. In order for an ensemble to outperform any of its individual components, the individual learners must be accurate and diverse enough to effectively capture the structure of the data (Hansen and Salamon 1990). However, determining the diversities of models to include is one challenging part of constructing an optimal ensemble. For the 2017 KDD cup, the winning team utilized an ensemble of 13 models including trees, neural networks and linear models (Hu et al. 2017). This diversity in the base learners is where the strength of an ensemble lies. Specifically, trees and neural networks are nonlinear models, where they partition the data space differently than linear models. As such, these models represent different features of the data, and once combined, can collectively represent the entire data space better than they would individually. However, in addition to determining the base models to be included there are two additional components that must be addressed. The first is how to tune the hyperparameters of each base model and the second is how to weight the base models to make the final predictions.

As previously stated, the construction of an ensemble model is a systematic process of combining many diverse base predictive learners. When aggregating predictive learners, there is always the question of how to weight each model as well as how to tune the parameters of the individual learners. One area that has not been given much attention is *how* to optimally tune hyperparameters of the diverse base models to obtain a better-performing ensemble model. The most straightforward approach is simply to average the pre-tuned base models, that is, all base models are given equal weight. However, numerous studies have shown that a simple average of models is not always the best and that a weighted ensemble can provide superior prediction results (Bhasuran 2016; Ekbal and Saha 2013; Winham et al. 2013; Peykani et al. 2019; Shahhosseini et al. 2019). Moreover, the hyperparameter tuning process for each base model is often carried out separately as an independent procedure when in fact it should be part of the training/learning framework. That is, implementations of a weighted ensemble consider the tuning of hyperparameters and weighting of models as two independent steps instead of as an integrated process. These gaps in the ensemble modeling serve as the major motivations for this study.

In this paper, we design an admissible framework for creating an optimal ensemble by considering the tuning of hyperparameters and weighting of models concurrently, something that is not previously considered by others. We implement a nested algorithm that is able to fill the gaps of finding optimal weights and tuning hyperparameters of ensembles in the literature. Moreover, we speed-up the learning and optimizing procedures by using a heuristic method based on Bayesian search instead of exhaustive search methods like grid search. For the traditional weighted ensemble creation methods, the hyperparameters are optimally tuned and they consider the tuning of hyperparameters and weights as independent processes, while this study's methodology does both at the same time and may select individually-non-optimal hyperparameters to create best ensembles.

To evaluate the designed algorithm, numerical experiments on several data sets from different areas have been conducted to demonstrate the generalizability of the designed scheme.

The main questions that we want to address in this paper are:

1) Does the designed method improve the diverse base learners?
2) How does the designed method compare to state-of-art ensemble techniques?
3) What is the effect of tuning hyperparameters as part of finding optimal ensemble weights on the quality of predictions?
4) Can the results be generalized to multiple data sets?

The remainder of this paper is organized as follows. Section 2 reviews the literature in the related fields; mathematics and concepts of the optimization model is presented in Section 3; the designed scheme (GEM-ITH) is introduced in Section 4; the results of comparing the designed method with benchmarks are presented and discussed in Section 5; and finally, Section 6 concludes the paper with major findings and discussions.

## 2. Background

A learning program is given data in the form $D = \{(X_i, y_i): X_i \in \mathbb{R}^{n \times p}, y_i \in \mathbb{R}\}$ with some unknown underlying function $y = f(x)$ where the $x_i$'s are predictor variables and the $y_i$'s are the responses with $n$ instances and $p$ predictor variables. Given a subset $S$ of $D$, a predictive learner is constructed on $S$, and given new values of $X$ and $Y$ not in $S$, predictions will be made for a corresponding $Y$. These predictions can be computed from any machine learning method or statistical model such as linear regression, trees or neural networks (Large et al. 2019). In the case where $Y$ is discrete, the learning program is a classification problem. If $Y$ is continuous, the learning program is a regression problem. The focus of this paper is on regression where the goal is to accurately predict continuous responses.

There have been extensive studies on weighted ensembles in the literature. The proposed approaches can be divided into constant and dynamic weighting. Perrone and Cooper (1992) presented two ensembling techniques in the neural networks' community. Basic Ensemble Method (BEM) combines several regression base learners by averaging their estimates. They demonstrate that BEM can reduce mean square error of the predictions by a factor of $N$, number of estimators. Moreover, Generalized Ensemble Method (GEM) was presented as the linear combination of the regression base learners and it was claimed that this ensemble method will avoid overfitting the data. The authors used cross-validation to make use of all training data in order to construct the ensemble estimators. Soon after, Krogh and Vedelsby (1995) proposed an optimization model to find the optimal weights of combining an ensemble of $N$ networks. They constrained the weights to be positive and sum to one in order to formulate generalization error and ambiguity of the ensemble to subsequently explain the bias-variance tradeoff using them. In addition, this study showed the importance of diversity and as they put it "it is important for generalization that the individuals disagree as much as possible". Another approach for constant ensemble weighting was using linear regression for finding the weights which was referred as stacked regression. This approach is similar to GEM, with a difference that the weights are not constrained to sum to one (Breiman 1996b). Another proposed method to combine base learners to build a better-performing ensemble is multi-stage neural network. In this method a second level of neural network estimator is trained on the first level base neural networks to create the ensemble (Yang and Browne 2004). It is obvious that the base first level learners can be any combination of machine learning models. Pham and Olafsson (2019a) proposed using the method of Cesaro averages for their weighting scheme essentially following a weighting pattern in line with Riemann zeta function with another generalization in Pham and Olafsson (2019b).

In the dynamic weighting approaches, the weights are assigned to each of the base learners according to their performance on the validation set. Jimenez and Walsh (1998) suggested a framework of dynamically averaging weights of a population of neural network estimators instead of using static performance-based weights. They formulated the prediction certainty and came up with a method to dynamically compute ensemble weights based on the certainty level each time the ensemble output was evaluated. The experimental results showed that the proposed methodology performed at least as well as the other ensemble methods and provided minor improvements in some cases. Shen and Kong (2004) proposed another dynamically weighted ensemble of neural networks for regression problems using the natural idea that higher training accuracy results in higher weight for a model.

Moreover, the applications areas in which ensemble approaches are used span a variety of areas. Belayneh et al. (2016) constructed an ensemble of bootstrapped artificial neural networks to predict drought conditions of a river basis in Ethiopia, whereas Martelli et al. (2003) constructed an ensemble of neural networks to predict membrane protein achieving superior results than previous methods. Aside from neural networks, Van Rijn et al. (2018) investigated the use of heterogeneous ensembles for data streams and introduced an online estimation framework to dynamically update the prediction weights of base learners. Zhang and Mahadevan (2019) constructed an ensemble of support vector machines to model the incident rates in aviation. Conroy et al. (2016) proposed a dynamic ensemble approach for imputing missing data in classification problems and compared the results of their proposed method with other common missing data approaches. A multi-target regression problem was addressed in a study by Breskvar et al. (2018) where ensembles of generalized decision trees with added randomization were used. Large et al. (2019) introduced a probabilistic ensemble weighting scheme based on cross-validation for classification problems. As evidenced in the literature, constructing an ensemble of models has many real-world applications due to the potential to achieve superior performance to that of a single model.

It can be observed that the existing ensembling studies all consider the base model construction and the weighted averaging to be independent steps. Intuitions tell us that considering the tuning of model parameters in conjunction with the weighted average should produce a superior ensemble. This intuition can be thought of in terms of the bias-variance tradeoff (Yu et al. 2006). Namely, if each base model is optimally tuned individually, then by definition they will have low bias but will have high variance. Therefore, by further combining these optimally tuned models we will create an ensemble that ultimately has low bias and high variance. However, by considering the model tuning and weighting as two concurrent processes (as opposed to independent), then we can balance both bias and variance to obtain an optimal ensemble – the goal of this paper. In this study, we designed a method that integrates the parameter tuning of the individual models and the ensemble weights design where the bias and variance trade-off is considered altogether in one decision-making framework.

To the best of our knowledge, there have not been studies that combine the model hyperparameter tuning and the model weights aggregation for optimal ensemble design in one coherent process. Motivated by this gap in the literature, we implement a nested optimization approach using cross-validation that accounts for optimizing hyperparameters and ensemble weights in different levels to address this issue. We formulated our model with the objective to minimize the prediction's mean squared error and account for the model hyperparameters and aggregate weights for each diverse predictive learner with a nonlinear convex program to find the best possible solution to the objective function from the considered search space.

## 3. Materials and methods

Ensemble learning has been shown to outperform individual base models in various studies (Perrone and Cooper 1992; Krogh and Vedelsby 1995; Brown 2017), but as mentioned previously, designing a systematic method to combine base models is of great importance. Based on many data science competitions, the winners are the ones who achieved superior performance by finding the best way to integrate the merits of different models (Puurula et al. 2014; Hong et al. 2014; Hoch 2015; Wang et al. 2015; Zou et al. 2017, Peykani et al. 2018). It has been shown that the optimal choice of weights aims to obtain the best prediction error by designing the ensembles for the best bias and variance balance (Krogh and Vedelsby 1995; Shahhosseini et al. 2020).

Prediction error of a model includes two components: bias and variance. Both are determined by the interactions between the data and model choice. Bias is a model's understanding of the underlying relationship between features and target outputs; whereas, variance is the sensitivity to perturbations in training data. For a given data set $D = \{(X_i, y_i): X_i \in \mathbb{R}^{n \times p}, y_i \in \mathbb{R}\}$, we assume there exists a function $f: \mathbb{R}^{n \times p} \to \mathbb{R}$ with noise $\epsilon$ such that $y = f(x_i) + \epsilon$ where $\epsilon \sim N(0,1)$.

Assuming the prediction of a base learner for the underlying function $f(x)$ to be $\hat{f}(x)$, We define bias and variance as follows.

$$Bias\ [\hat{f}(x)] = E[\hat{f}(x)] - f(x) \qquad [1]$$

$$Var[\hat{f}(x)] = E[\hat{f}(x)^2] - E[\hat{f}(x)]^2 \qquad [2]$$

Based on bias-variance decomposition (Hastie et al. 2005) the above definitions for bias and variance can be aggregated to the following:

$$E\left[\left(f(x) - \hat{f}(x)\right)^2\right] = \left(Bias\left[\hat{f}(x)\right]\right)^2 + Var[\hat{f}(x)] + Var(\epsilon) \qquad [3]$$

The third term, $Var(\epsilon)$, in Equation [3] is called irreducible error, which is the variance of the noise term in the true underlying function ($f(x)$) and cannot be reduced by any model (Hastie et al. 2005).

The learning objective of every prediction task is to approximate the true underlying function with a predictive model that has low bias and low variance, but this is not always accessible. Common approaches to reduce variance are cross-validation and bagging (bootstrapped aggregated ensemble). On the other hand, reducing bias is done commonly with boosting. Although each of these approaches has its own merits and shortcomings, finding the optimal balance between them is the main challenge (Zhang and Ma 2012).

To find the optimal way to combine base learners, a mathematical optimization approach is used that is able to find ensemble optimal weights. We consider regression problems that have continuous targets to predict in this article. Majorly taking prediction bias into account, and knowing that mean squared error (MSE) is defined as the expected prediction error ($E[(f(x) - \hat{f}(x))^2]$) (Hastie et al. 2005), the objective function in the mathematical model for optimizing ensemble weights is chosen to be MSE (Shahhosseini et al. 2020).

Moreover, as several studies have shown, using cross-validation to find optimal weights is effective in reducing the variance to some extent. The smoothing property of ensemble estimators which is defined as the ability of the ensemble model to make use of regression ensembles coming from different sources, alleviates the over-fitting problem (Perrone and Cooper 1992). In addition, to ensure the base learners are diverse, it makes sense to train them on different training sets using cross-validation procedures, as well as selecting diverse estimators as base learners (Krogh and Vedelsby 1995).

The following optimization model (GEM) which was proposed by Perrone and Cooper (1992) intends to find the best way to combine predictions of base learners by finding the optimal weight to aggregate them in a way that the created ensemble minimizes the total expected prediction error (MSE). Note that the out-of-bag predictions of each base learner ($\hat{Y}_i$) are the predictions of trained base learners on the hold-out set of an $m$-fold cross-validation.

$$\text{Min } MSE(w_1\hat{Y}_1 + w_2\hat{Y}_2 + \cdots + w_k\hat{Y}_k, Y) \qquad [4]$$
$$s.t.$$
$$\sum_{j=1}^{k} w_j = 1,$$
$$w_j \geq 0, \quad \forall j = 1, \ldots, k.$$

where $w_j$ is the weights corresponding to base model $j$ ($j = 1, \ldots, k$), $\hat{Y}_j$ represents the vector of out-of-bag predictions of base model $j$ on the validation instances of cross-validation, and $Y$ is the vector of true response values. Assuming $n$ is the total number of instances, $y_i$ as the true value of observation $i$, and $\hat{y}_{ij}$ as the prediction of observation $i$ by base model $j$, the optimization model is as follows.

$$Min \ \frac{1}{n}\sum_{i=1}^{n}\left(y_i - \sum_{j=1}^{k} w_j \hat{y}_{ij}\right)^2 \quad\quad\quad [5]$$
$$s.t.$$
$$\sum_{j=1}^{k} w_j = 1,$$
$$w_j \geq 0, \quad \forall j = 1, \dots, k.$$

The above formulation is a nonlinear convex optimization problem. As the constraints are linear, computing the Hessian matrix will demonstrate the convexity of the objective function. Hence, since a local optimum of a convex function (objective function) on a convex feasible region (feasible region of the above formulation) is guaranteed to be a global optimum, the optimal solution of this problem is proved to be the global optimal solution (Boyd and Vandenberghe 2004).

The GEM algorithm is displayed below.

---

**Inputs:** Data set $D = \{(x, y): x \in \mathbb{R}^{n \times p}, y \in \mathbb{R}^n\}$;
$k$ base learning algorithm;
For $j = 1, \dots, k$:
    For $i = 1, \dots, m$ splits:      % $m$-fold cross-validation
        Split $D$ into $D_i^{train}, D_i^{test}$ for the $i$th split
        Train base learner $j$ on $D_i^{train}$
        $P_{ij}$: Predict on $D_i^{test}$
    End.
    $\hat{Y}_j = (P_{1j}, \dots, P_{mj})$      % Concatenate $m$ predictions on $D_i^{test}$
End.
Use $\hat{Y}_j$'s to Compute $w_j$ from optimization problem [4]
Combine base learners $1, \dots, k$ with weights $w_1, \dots, w_k$.
**Outputs:** Optimal objective value ($MSE^*$)
        Optimal ensemble weights ($w_1^*, \dots, w_k^*$)
        Predictions of the ensemble with optimal weights ($\hat{Y}^*$)

---

*The Generalized Ensemble Model (GEM) algorithm*
*The input data set is $D = \{(x, y): x \in \mathbb{R}^{n \times p}, y \in \mathbb{R}^n\}$. $k$ base learners are considered as input base learners. $m$-fold cross-validation is used to generate out-of-bag predictions which are the inputs to the optimization model ($\hat{Y}_j$). The optimal weights ($w_j^*$) are used to combine $j$ base learners and make final predictions ($\hat{Y}^*$).*

The Generalized Ensemble Model (GEM) assumes hyperparameters of each base learner is tuned before conducting the ensemble weighting task. For example, if one of the base learners is the random forest, its hyperparameters are tuned with one of the many common tuning approaches and the predictions made with the tuned model act as the inputs of the optimization model to find the optimal ensemble weights. One of the main questions of this study is whether the best performing ensemble results from the set of tuned hyperparameters. To answer this question, an algorithm is designed which is based on optimization. This algorithm makes it possible to find the best set of hyperparameters from the considered search space, that results in the best-performing ensemble.

## 4. Generalized Ensemble Model with Internally Tuned hyperparameters (GEM-ITH)

Generalized Ensemble Model (GEM), which is a nonlinear optimization model was presented in section 3 to find the optimal weights of combining different base learner predictions. In this section, we want to investigate the effect of tuning hyperparameters of each base learner on the optimal ensemble weights. A common approach in creating ensembles is tuning hyperparameters of each base model with different searching methods like grid search, random search, Bayesian optimization, etc., independently and then combine the predictions of those tuned base learners by some weights. We claim here that the ensemble with the best prediction accuracy (the least mean squared error) may not be created from hyperparameters tuned individually. To this end, we have designed an optimization based nested algorithm that aims to find the best combination of hyperparameters from the considered combinations that results in the least prediction error. Fig.1 demonstrates a flow chart of traditional weighted ensemble creation (GEM) and GEM–ITH, respectively.

The designed nested algorithm can find the best optimal solution from the considered search space when using greedy search methods such as grid search. However, in that case, performing optimization task may not be efficient. Therefore, to speed-up this process we make use of a heuristic based on Bayesian search that aims at finding some candidate hyperparameter values for each base learner and obtain the best weights and hyperparameters combination for the ensemble of all base models. Although the best weights and hyperparameters found by this heuristic are not necessarily as good as best combinations found by grid search, they approach those values after enough iterations.

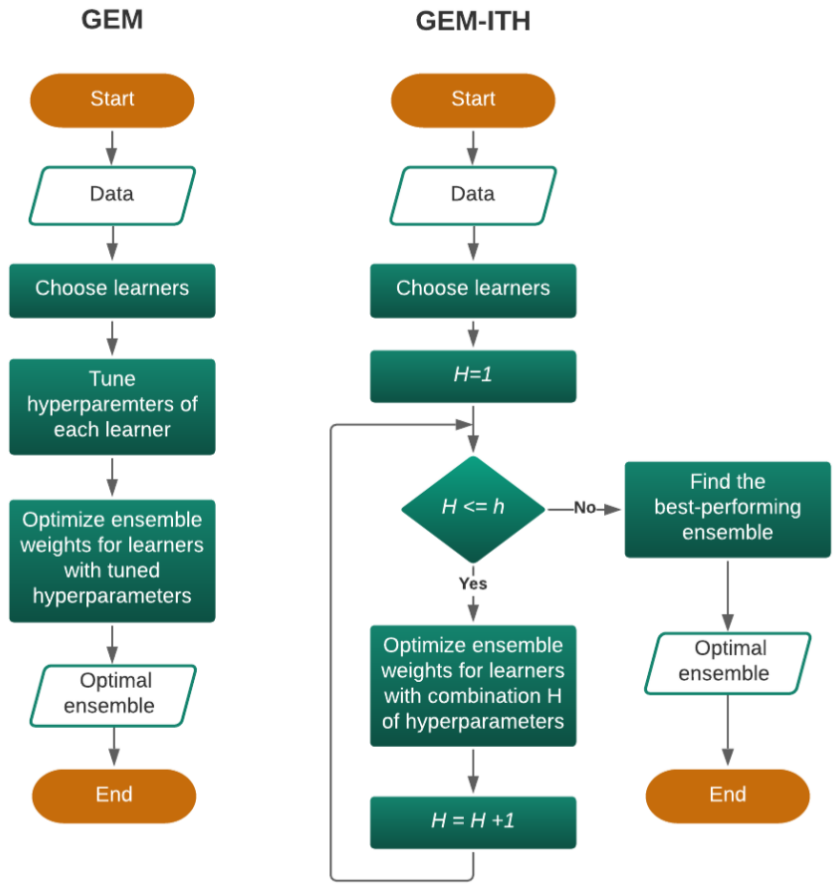

*Fig.1 traditional weighted ensemble creation flowchart (GEM) vs. GEM-ITH flowchart. For the GEM ensemble creation methods, the hyperparameters are optimally tuned as an independent process. The GEM-ITH method searches across all hyperparameter combinations of $k$ base learners ($h = |h_1| \times ... \times |h_k|$ when $h_j$ is the set of all hyperparameter combinations of model $j$).*

### 4.1. GEM-ITH with Bayesian search

Bayesian optimization aims to approximate the unknown function with surrogate models like Gaussian process. The main difference between Bayesian optimization and other search methods is incorporating prior belief about the underlying function and updating it with new observations. Bayesian optimization tries to gather observations with the highest information in each iteration by making a balance between exploration (exploring uncertain hyperparameters) and exploitation (gathering observations from hyperparameters close to the optimum) (Snoek et al. 2012).

Given $b$ iterations of Bayesian optimization, $b$ hyperparameter combinations for each base learner have been identified resulting in $b^k$ total number of combinations that should be considered by GEM-ITH model. Each of these combinations in turn is used to calculate out-of-bag predictions of each base learner and treat them as the inputs to the optimization model [4].

---

**Inputs:** Data set $D = \{(x, y): x \in \mathbb{R}^{n \times p}, y \in \mathbb{R}^n\}$;
$k$ base learning algorithm;
Hyperparameters sets $h_1, \ldots, h_k$

Bayesian search chooses $b$ hyperparameter combination for each learner

For $H = 1, \ldots, b^k$:
    For $j = 1, \ldots, k$:
        For $i = 1, \ldots, m$ splits:     % $m$-fold cross-validation
           Split $D$ into $D_i^{train}, D_i^{test}$ for the $i$th split
           Train base learner $j$ with hyperparameter combination $H$
              on $D_i^{train}$
           $P_{ij}$: Predict on $D_i^{test}$
        End.
        $\hat{Y}_j = (P_{1j}, \ldots, P_{mj})$     % Concatenate $m$ predictions on $D_i^{test}$
    End.
    Use $\hat{Y}_j$'s to Compute $w_j$ from optimization problem [4]
    Calculate optimal objective value ($MSE_H^*$), optimal weights, $(w_{1H}^*, \ldots, w_{kH}^*)$, and ensemble predictions ($\hat{Y}_H^*$)
End.

Find the minimum of objective values ($MSE_H^*$).
Find the optimal weights $w_{1H}^*, \ldots, w_{kH}^*$ corresponding to the minimum objective value.

**Outputs:** Optimal objective value ($MSE^*$)
Optimal combination of hyperparameters $h_1^*, h_2^*, \ldots, h_p^*$.
Optimal ensemble weights $w_1^*, \ldots, w_k^*$
Prediction vector of ensemble with optimal weights ($\hat{Y}^*$)

---

*The GEM-ITH algorithm with Bayesian search.*
*From the input hyperparameter sets of each base learner, Bayesian search selects $b$ combinations, resulting in a total of $b^k$ combinations. For each combination, predictions on the hold-out sets of cross-validation are used to find the optimal ensemble weights and objective value. The best hyperparameter combination and optimal solution is selected by finding the one with the minimum objective value.*

## 5. Results and discussion

### 5.1. Numerical experiments

To evaluate the designed algorithm, numerical experiments on multiple data sets from UCI Machine Learning Repository[1] (Dua and Graff 2019), Scikit learn data sets (Pedregosa et al.

---

[1] https://archive.ics.uci.edu/ml/index.php

2011), and Kaggle data sets from a variety of domains have been conducted to demonstrate the generalizability of the designed scheme. Details of these data sets are shown in Table.1 (Ferreira et al. 2010; Yeh 1998; Efron et al. 2004; Arzamasov et al. 2018; Tsanas and Xifara 2012; Acharya et al. 2019; Grisoni et al. 2016; Cassotti et al. 2015; Cortez et al. 2009)

*Table.1 data sets chosen to evaluate GEM-ITH*

|   | Data sets | Number of Instances | Number of Attributes | Area |
|---|---|---|---|---|
| 1 | Behavior of Urban Traffic of Sao Paolo | 135 | 18 | Computer |
| 2 | Concrete Compressive Strength | 1030 | 9 | Physical |
| 3 | Diabetes Data | 442 | 10 | Life |
| 4 | Electrical Grid Stability Simulated Data | 10000 | 14 | Physical |
| 5 | Energy efficiency | 768 | 8 | Computer |
| 6 | Graduate Admissions | 500 | 9 | Education |
| 7 | QSAR Bioconcentration Classes | 779 | 14 | Life |
| 8 | QSAR Fish Toxicity Data | 908 | 7 | Physical |
| 9 | Wine Quality | 4898 | 12 | Business |
| 10 | Yacht Hydrodynamics | 308 | 7 | Physical |

Four machine learning algorithms with minimal pre-processing tasks were designed for each data set separately and the designed algorithm is applied to them. Five-fold cross-validation was used for generating out-of-bag predictions for all designed ML models and the entire process was repeated 5 times. In addition, 20% of each data set was reserved for testing and the training and optimizing procedure was done on the remaining 80%.

## 5.2. Base models generation

A heuristic method was used here to generate base learners. Two important aspects of the base learners were considered in this heuristic: 1) diversity, 2) performance. We intended to select four base learners that show a certain level of diversity and performance to eventually create a well-performing ensemble model. The following steps were taken to generate base learners.

1) *Trial training*: Many machine learning models were trained on each of the considered data sets and their performance were evaluated using unseen test observations (See Table.2 for the hyperparameter settings of the models).
2) *Performance pruning*: The trained models whose prediction error were higher than the average prediction error of all trained models, were removed from the pool of the initial models.
3) *Correlation*: Pair-wise correlation of the remaining models were calculated.
4) *Rank*: The pair-wise correlations were ranked from the lowest correlation to the highest.
5) *Selection*: The top four models with the least pair-wise correlations were selected as the base models for ensemble creation.

*Table.2 Initial ML models and their hyperparameters settings*
*All models were trained using scikit learn package*

| ML Model | Hyperparameter | Values |
|---|---|---|
| Ridge | alpha | 10^range(-5, 0)[2] |
| LASSO | alpha | 10^range(-5, 0) |
| Elastic Net | alpha | 10^range(-5, 0) |
|  | l1_ratio | 10^range(-5, 0) |
| LARS | n_nonzero_coefs | range(1,p-1)[3] |
| Orthogonal Matching Pursuit | n_nonzero_coefs | range(1,p-1) |
| Bayesian Ridge | alpha_1 | 10^range(-5, 0) |
|  | alpha_2 | 10^range(-5, 0) |
| SGD Regressor | alpha | 10^range(-5, 0) |
|  | l1_ratio | 10^range(-5, 0) |
| SVM | C | linspace(0.01, 5, 20)[4] |
|  | gamma | range(0.01, 0.5, 0.05) |
|  | kernel | {linear, poly, rbf} |
| KNN | n_neighbors | range(2,11) |
| Gaussian Process Regressor | alpha | 10^range(-10, -5) |
| Regression tree | max_depth | range(4,23) |
| Bagging | n_estimators | {100, 200, 500} |
|  | max_samples | {0.7, 0.8, 0.9, 1.0} |
| Random Forest | n_estimators | {100, 200, 500} |
|  | max_depth | range(4,10) |
| Extremely Randomized Trees | n_estimators | {100, 200, 500} |
|  | max_depth | range(4,10) |
| AdaBoost Regressor | n_estimators | {100, 200, 500} |
|  | learning_rate | linspace(0.5, 2, 20) |
| Gradient Boosting Regressor | n_estimators | {100, 200, 500} |
|  | learning_rate | linspace(0.5, 2, 20) |
| XGBoost | gamma | {5, 10} |
|  | learning_rate | {0.1, 0.3, 0.5} |
|  | n_estimators | {50, 100, 150} |
|  | max_depth | {3, 6, 9} |
|  | gamma | range(0.01, 0.55, 0.05) |
| Neural network | alpha | linspace(0.0001, 0.5, 20) |
|  | learning_rate_init | linspace(0.0001, 0.5, 20) |
|  | activation | {identity, logistic, tanh, relu} |

In the ideal implementation of the GEM-ITH method, grid search can be used to find the optimal hyperparameters and ensemble weights. However, since that is computationally expensive and difficult to implement in practice, we use Bayesian search to find top 12 combinations of the hyperparameters of each ML model. Therefore, since we select four ML models with the heuristic explained above, the model should consider $12^4$ combination of ML models hyperparameters. It should be noted that uniform settings have been selected for Bayesian search. In other words, Bayesian search looks through all uniform values in the range of hyperparameters. All other ensemble models and base learners are trained using discrete settings of grid search.

---

[2] Numbers between 10^(-5) and 1
[3] Numbers between 1 and p-1 (p is the number of predictor variables)
[4] 20 linearly spaced numbers between 0.01 and 5

To conduct the Bayesian search *hyperopt* package (Bergstra et al. 2013) was used in Python 3. Also, Sequential Least Squares Programming algorithm (SLSQP) from Python's SciPy optimization library were used to solve optimization problems (Jones et al. 2001)

### 5.3. Benchmarks

Apart from the Generalized Ensemble Method introduced in Section 3 (GEM), four other state-of-art benchmarks have been used to compare the results of the designed learning methodology with them.

1) The first benchmark is the Generalized Ensemble Method introduced in Section 3 (GEM),
2) The second benchmark is the ensembles constructed with averaging the input base models (BEM).
3) Stacked ensemble with linear regression as the second level learner serves as the third benchmark, which we call stacked regression. This benchmark has been widely used as one of the most effective methods to create ensembles and is created with fitting a linear regression model on the predictions made by different base learners (Clarke 2003; Yao et al. 2018; Matlock 2018; Pavlyshenko 2019).
4) Considering random forest as one of the most powerful machine learning models as the second level of stacking, we construct stacked ensemble with random forest as the fourth benchmark (Thøgersen et al. 2016; Zhang et al. 2018).
5) Lastly, Stacked ensemble with *k*-nearest neighbor model as the 2nd level training model is added as the fifth state-of-art benchmark (Ozay and Yarman-Vural 2016; Pakrashi and Mac Namee 2017).

### 5.4. Numerical results

Table.3 shows the average results of GEM-ITH based off of Bayesian search methods along with mean squared error of predictions made by each base learner and benchmarks. The superiority of the designed ensemble techniques can be seen by comparing their prediction errors with base learners. This answers the first question asked in the Introduction section and demonstrates the improvements of the GEM-ITH over base learners.

*Table.3 The average results of applying ML models and created ensembles on 10 public data sets*
*Base models (Models 1 to 4) are different for different data sets and are generated using a heuristic. The best prediction accuracy (lowest prediction error) in each row is shown in bold*

| Data set | Objective value on test set (MSE) | | | | | | | | | |
|---|---|---|---|---|---|---|---|---|---|---|
| | Model 1 | Model 2 | Model 3 | Model 4 | BEM | Stacked Regression | Stacked RF | Stacked KNN | GEM | GEM-ITH |
| Behavior of Urban Traffic | 8.42 | 7.63 | 7.83 | 7.46 | 7.10 | 7.69 | 7.85 | 7.30 | 7.67 | **7.06** |
| Concrete Compressive Strength | 39.02 | 19.53 | 23.36 | 28.94 | 19.44 | 18.85 | 23.32 | 24.04 | 19.12 | **18.61** |
| Diabetes Data | 3042.27 | 3066.53 | 3110.75 | 5165.84 | 3122.05 | 3055.35 | 3884.18 | 3572.87 | 3038.89 | **2987.23** |
| Electrical Grid Stability ($\times 10^4$) | 3.55 | 2.79 | 13.58 | 4.70 | 3.70 | 1.82 | 2.14 | **2.12** | 2.36 | 2.25 |
| Energy efficiency | 4.06 | 10.63 | 1.62 | 11.64 | 4.49 | 1.45 | 2.01 | 2.12 | 1.62 | **1.42** |
| Graduate Admissions ($\times 10^3$) | 3.63 | 4.26 | 19.74 | 4.62 | 5.01 | 3.58 | 4.31 | 4.22 | 3.60 | **3.52** |
| QSAR Bioconcentration ($\times 10$) | 6.69 | 5.54 | 5.83 | 5.59 | 5.51 | 5.36 | 6.55 | 6.09 | 5.34 | **5.27** |
| QSAR Fish Toxicity | 8.51 | 7.67 | 7.68 | 7.03 | 7.05 | 7.09 | 9.27 | 8.78 | 7.04 | **6.93** |
| Wine Quality ($\times 10$) | 4.49 | 4.55 | 4.24 | 3.64 | 4.01 | 3.63 | 4.23 | 4.27 | 3.64 | **3.62** |
| Yacht Hydrodynamics | 70.93 | 1.15 | 0.88 | 69.42 | 15.55 | 0.96 | 1.66 | 1.61 | 0.91 | **0.77** |

Table.4 demonstrates the different choices of hyperparameters as the optimal selections for creating optimal ensembles from GEM and GEM-ITH for Energy Efficiency data set (the same was observed for other data sets, but they are not shown here). Comparing the tuned hyperparameters before creating ensembles, with the ones found by GEM-ITH, the main claim of this paper is proved to be true. The hyperparameters found to be optimal by GEM-ITH method are different from the hyperparameters tuned separately (GEM). This means that in order to create better performing ensembles, the hyperparameters should not necessarily be the ones that are proved to be the best independently. This addresses the third question from questions raised in the introduction section and expresses that tuning hyperparameters as part of finding optimal ensemble weights results in higher quality predictions.

*Table.4 Comparing optimal hyperparameters of GEM and GEM-ITH for Energy Efficiency data set*

| Hyperparameter | Ensemble Method | Hyperparameter value |
|---|---|---|
| Regression Tree (*max_depth*) | GEM | 6 |
| | GEM-ITH | 19 |
| Elastic Net (alpha) | GEM | 0.00001 |
| | GEM-ITH | 0.76785 |
| Elastic Net (l1_ratio) | GEM | 0.00001 |
| | GEM-ITH | 0.01317 |
| XGBoost (*gamma*) | GEM | 5 |
| | GEM-ITH | 6.92567 |
| XGBoost (learning_rate) | GEM | 0.1 |
| | GEM-ITH | 0.41613 |
| XGBoost (n_estimators) | GEM | 150 |
| | GEM-ITH | 150 |
| XGBoost (*max_depth*) | GEM | 9 |
| | GEM-ITH | 9 |
| SVM (*C*) | GEM | 1.32315 |
| | GEM-ITH | 4.92209 |
| SVM (*gamma*) | GEM | 0.01 |
| | GEM-ITH | 0.35520 |

Fig.2 exhibits the normalized error rates of data sets under study for the designed ensemble models. It visualizes the comparison between GEM-ITH and the state-of-art benchmarks. The figure shows almost complete dominance of GEM-ITH over the benchmarks addressing the second question raised in the introduction section. GEM-ITH has been the winner in 9 out of 10 public data sets.

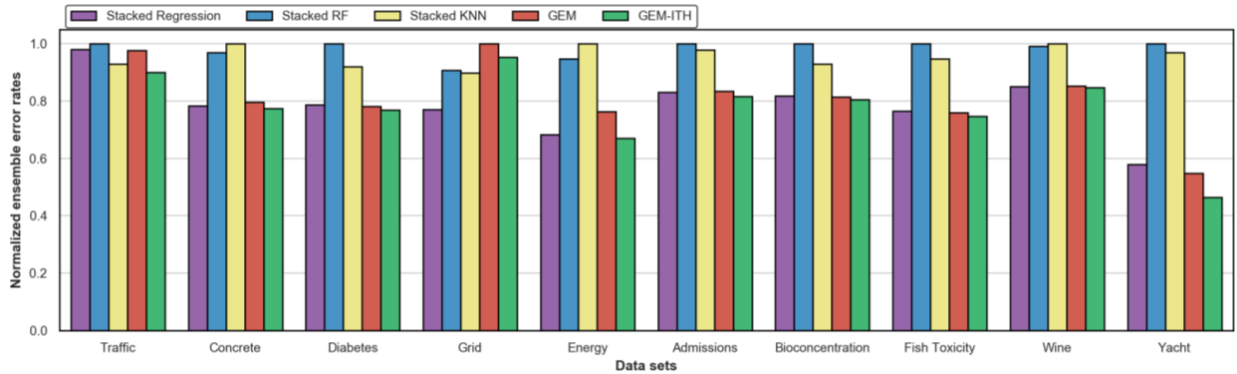

*Fig.2 Comparison of the GEM-ITH vs. state-of-art benchmarks on 10 public data sets*
*(average normalized test errors)*

Hence, it can be concluded that the designed scheme (GEM-ITH) improves the prediction accuracy of each base learner. Comparing them to the state-of-art ensemble methods, GEM-ITH could achieve better prediction accuracy among all, while introducing an improvement over successful GEM scheme. Therefore, this confirms the hypothesis that tuning hyperparameters of base learners inside optimal ensemble creating procedure will result in better prediction accuracy. These findings demonstrate the generalizability of GEM-ITH to real data sets since we have applied the methods on 10 publicly available data sets with diverse properties, which addresses the last question raised at the end of the Introduction section and shows the generalizability of the designed method on multiple data sets.

All the models have been run on a computer equipped with a 2.6 GHz Intel E5-2640 v3 CPU, and 128 GB of RAM. The computation time of each model is shown in the Table.5. The computation time depends heavily on the complexity of the selected base learners and the dimensions of the data set. All in all, due to the high complexity of the designed model (GEM-ITH), it is more appropriate to be used for small to medium size data sets.

Table.5 Computation time of all trained models for each data set

| Data set | Computation time (seconds) | | | | | | | | | |
|---|---|---|---|---|---|---|---|---|---|---|
| | Model 1 | Model 2 | Model 3 | Model 4 | BEM | Stacked Regression | Stacked RF | Stacked KNN | GEM | GEM-ITH |
| Behavior of Urban Traffic | 0.65 | 18.06 | 6.28 | 6.61 | 162.18 | 162.18 | 162.19 | 162.18 | 162.50 | 7862.99 |
| Concrete Compressive Strength | 0.46 | 18.80 | 39.88 | 1393.58 | 8113.88 | 8113.92 | 8113.94 | 8113.92 | 8114.35 | 26387.18 |
| Diabetes Data | 0.12 | 20.13 | 0.64 | 15.72 | 232.53 | 232.54 | 232.55 | 232.54 | 232.91 | 11143.46 |
| Electrical Grid Stability | 6.12 | 20.20 | 291.22 | 1.04 | 1572.30 | 1572.43 | 1572.62 | 1572.44 | 1572.76 | 19739.76 |
| Energy efficiency | 0.40 | 47.14 | 11.89 | 68.59 | 462.39 | 462.41 | 462.42 | 462.41 | 462.54 | 8575.98 |
| Graduate Admissions | 0.08 | 0.42 | 9.92 | 13.13 | 95.72 | 95.74 | 95.75 | 95.74 | 96.32 | 12999.11 |
| QSAR Bioconcentration | 0.66 | 31.22 | 13.70 | 34.93 | 386.30 | 386.32 | 386.34 | 386.32 | 386.81 | 12783.72 |
| QSAR Fish Toxicity | 0.64 | 34.26 | 15.75 | 72.25 | 576.46 | 576.47 | 576.49 | 576.47 | 576.98 | 13788.44 |
| Wine Quality | 0.17 | 0.89 | 33.31 | 28.12 | 555.86 | 555.91 | 555.93 | 555.91 | 556.42 | 37854.14 |
| Yacht Hydrodynamics | 0.29 | 9.83 | 6.34 | 13.92 | 183.46 | 183.52 | 183.53 | 183.52 | 183.71 | 26300.61 |

# 6. Conclusion

In an attempt to observe the effect of tuning hyperparameters of base learners on the created ensembles, an optimization based nested algorithm that finds the optimal weights to combine base learners as well as the optimal set of hyperparameters for each of them (GEM-ITH) was designed in this study. To address the complexity issues, Bayesian search was used to generate base learners and a heuristic algorithm was used to generate base learners that exhibit a certain level of diversity and performance. The designed methods were applied to ten public data sets and compared to state-of-art ensemble techniques. Based on the obtained results, it was shown that GEM-ITH is able to dominate state-of-art ensemble creation methods. Furthermore, it was demonstrated that the hyperparameters used in creating optimal ensembles are different when they are tuned internally with GEM-ITH algorithm, than when they are tuned independently (GEM).

This study is subject to a few limitations, which suggest future research directions. Firstly, designing a nested algorithm for classification problems could expand the algorithm to classification problems and investigate its effectiveness on them. Secondly, applying a similar concept of hyperparameter tuning on other ensemble creating methods such as regularized stacking will more demonstrate the impact of hyperparameter tuning when creating ensembles. Lastly, trying to speed-up the ensemble creating process even more when considering hyperparameter tuning will create a competitive edge for the algorithm over competitions.